\documentclass[a4paper]{article}

\usepackage{multirow}

\usepackage{booktabs}
\usepackage{siunitx}
\usepackage{graphicx}
\usepackage{verbatim}

\usepackage{INTERSPEECH2020}

\title{Are Neural Open-Domain Dialog Systems Robust to Speech Recognition Errors in the Dialog History? An Empirical Study}

\name{Karthik Gopalakrishnan, Behnam Hedayatnia, Longshaokan Wang, Yang Liu, Dilek Hakkani-T\"{u}r}
\address{Amazon Alexa AI}
\email{\{karthgop, behnam, longsha, yangliud, hakkanit\}@amazon.com}

\begin{document}

\maketitle
\begin{abstract}
  Large end-to-end neural open-domain chatbots are becoming increasingly popular. However, research on building such chatbots has typically assumed that the user input is written in nature and it is not clear whether these chatbots would seamlessly integrate with automatic speech recognition (ASR) models to serve the speech modality. We aim to bring attention to this important question by empirically studying the effects of various types of synthetic and actual ASR hypotheses in the dialog history on TransferTransfo, a state-of-the-art Generative Pre-trained Transformer (GPT) based neural open-domain dialog system from the NeurIPS ConvAI2 challenge. We observe that TransferTransfo trained on written data is very sensitive to such hypotheses introduced to the dialog history during inference time.
  As a baseline mitigation strategy, we introduce synthetic ASR hypotheses to the dialog history during training and observe marginal improvements, demonstrating the need for further research into techniques to make end-to-end open-domain chatbots fully speech-robust. To the best of our knowledge, this is the first study to evaluate the effects of synthetic and actual ASR hypotheses on a state-of-the-art neural open-domain dialog system and we hope it promotes speech-robustness as an evaluation criterion in open-domain dialog.
\end{abstract}
\noindent\textbf{Index Terms}: neural open-domain dialog systems, speech recognition, response generation, response selection

\section{Introduction}

Neural modeling approaches are prominent in research on both task-oriented and open-domain dialog. Traditional sequence-to-sequence models have been used for encoding the dialog history and predicting domains, intents, slot types, spans and more generally decoding full-fledged system responses. In recent years, large pre-trained Transformer-based models for natural language understanding (NLU) and
natural language generation (NLG) have become ubiquitous~\cite{devlin2018bert,radford2019language},
leading to tremendous advances by fine-tuning towards these dialog tasks~\cite{zhang2019find,zhang2019dialogpt}.

\begin{figure}[htb]
  \centering
\footnotesize
\begin{tabular}{ l l }
\toprule
\textbf{}
& \textbf{Message}
\\
\midrule
\dots & \dots \\
User & \begin{tabular}{@{}l@{}}\textit{That is insane, \textbf{but I heard that}} \textit{is \textbf{cost} like} \\ \textit{\textbf{\$110 million}, that is} \textit{a \textbf{ton of} money}\end{tabular} \\
\\
ASR Output & \begin{tabular}{@{}l@{}}\textit{that is insane \textbf{by her Ladic}} \textit{is \textbf{coast} like,} \\ \textit{\textbf{hungry and 10}} \textit{\textbf{medium}. That is a \textbf{tongue off}} \\ \textit{money.}\end{tabular} \\

\midrule

Written Text Model & \begin{tabular}{@{}l@{}}that is crazy, i wonder if she can eat the food?\end{tabular} \\
\midrule
ASR Robust Model & \begin{tabular}{@{}l@{}}yeah, it is. that is a lot of money. i think that \\ the american people need that much money.\end{tabular} \\
\bottomrule
\end{tabular}
\caption{\label{sample_conversation} {A snippet from our augmented Topical-Chat
audio test set, showing the ASR output for the user utterance in the dialog history
and responses by two models: i) Written Text Model and ii) ASR Robust Model. Written Text Model is thrown off by the ASR output and responds about food, whereas ASR Robust Model responds correctly about money.}}

\vspace{-0.15in}
\end{figure}

In task-oriented speech-based dialog systems, the effect of ASR hypotheses has been widely studied and techniques have been devised to minimize the resulting downstream NLU errors~\cite{yang2015using, masumura2018neural, khan2015hypotheses, shivakumar2019spoken, schumann2018incorporating, hakkani2006beyond, thomson2008evaluating}. More recently, end-to-end spoken language understanding
approaches have been attempted to sidestep this problem~\cite{haghani2018audio, serdyuk2018towards}.
On the other hand, research in open-domain dialog is increasingly focusing on large, monolithic end-to-end neural models like Google's Meena~\cite{adiwardana2020towards} that are built using written data and evaluated on written interactions. Several written textual datasets have been created recently~\cite{zhang2018personalizing,dinan2018wizard,gopalakrishnan2019topical} to tackle various problems in open-domain dialog, including persona-grounding, knowledge-grounding and reasoning, and state-of-the-art chatbots have been built using them. However, it is not clear whether these written text-based open-domain chatbots would seamlessly integrate with ASR models to serve the speech modality, which is popular due to the ubiquity of voice assistants like Alexa, Google Assistant and Siri.

Collecting large-scale written text-based dialog datasets is cheaper and more practical than collecting audio-based dialog datasets in many ways. But speech-robustness should be a factor of consideration when designing any (task-oriented or open-domain) dialog system intended to be deployed to the speech modality, even in the absence of audio-based training data. To bring attention to this important aspect in the open-domain dialog community, we empirically study the effects of various types of synthetic and actual ASR hypotheses in the dialog history on TransferTransfo (TF2)~\cite{wolf2019transfertransfo}, a state-of-the-art neural open-domain dialog system based on the Generative Pre-trained Transformer (GPT)~\cite{radford2018improving} from the NeurIPS ConvAI2 Conversational Intelligence Challenge~\cite{dinan2020second}.
We build off the Topical-Chat dataset~\cite{gopalakrishnan2019topical} and perform two augmentations in our study: one creating \textit{simulated} ASR hypotheses for the entire dataset, and another creating \textit{actual} ASR hypotheses with a smaller audio-based analogue of the Topical-Chat test sets.

We observe that TF2 trained on written textual data is very sensitive to synthetic and actual ASR hypotheses introduced to the dialog history during inference time, with the sensitivity being particularly prominent for the task of response selection.
As a baseline mitigation strategy, we introduce synthetic ASR hypotheses to the dialog history during training and observe marginal improvements, demonstrating the need for further research into techniques to make end-to-end open-domain chatbots fully speech-robust. Figure~\ref{sample_conversation} shows a sample snippet
with responses from TF2 models trained on written text and synthetic ASR hypotheses when fed speech-distorted dialog history.

A close work to ours in spirit is~\cite{sankar2019neural}, which shows that Transformer-based generative dialog models are insensitive to \textit{unrealistic} perturbations like token-shuffling of the dialog history. Our work is more focused on
evaluating the effects of introducing \textit{realistic} perturbations to the dialog history in the form of synthetic and actual ASR hypotheses. Our augmentation of Topical-Chat, dubbed the Topical-Chat ASR dataset, is open-sourced\footnote{https://github.com/alexa/Topical-Chat/tree/master/TopicalChatASR/} to enable open-domain dialog researchers to perform speech-robustness evaluation and fuel research into novel techniques to make monolithic neural open-domain dialog models more speech-robust.

\vspace{-0.05in}
\section{Preliminaries}

\subsection{Notation}
Let $D_t = [x_1, \dots, x_t]$ denote a dialog history containing a sequence of $t$ turns. Let $H_t = x_1 \oplus \dots \oplus x_t$ denote a flattened sequence of all tokens in $D_t$. Let $W$ be a truncate parameter for a flattened dialog history $H$, which retains at most $W$ tokens from the end in $H$. When applied to $H_t$, we denote it as $H_t^W$. Finally, we denote a dialog example pair by $(H_t^W, c_{t+1})$, where $c_{t+1}$ is a candidate response that is either the ground-truth response $x_{t+1}$ at turn $t+1$ or a distractor response.

\vspace{-0.05in}
\subsection{Data}


For our experiments, we use the dialogs from Topical-Chat~\cite{gopalakrishnan2019topical}. This is one of the largest and most diverse knowledge-grounded text-based open-domain dialog datasets publicly available today. Each dialog in Topical-Chat contains 20+ turns alternating between two Turkers and $\sim$19 tokens per turn on average. Topical-Chat has two types of test sets: \textit{test freq} and \textit{test rare}.

We performed two augmentations of Topical-Chat. First, a \textit{simulated} augmentation wherein simulated errors are introduced at a corpus-level target word error rate (WER). For each data split (train/valid/test), we simulated ASR hypotheses for each turn in a dialog with four WER settings (0.1, 0.15, 0.2 and 0.3), each with a single seed for train and five random seeds for valid and test. We used the ASR error simulator method based on $n$-gram confusion matrix~\cite{fazel2019investigation, schatzmann2007error} and trained the simulator on transcribed ASR output from an internal user study.

Second, an \textit{actual} augmentation wherein two new test sets were created: \textit{test freq audio} and \textit{test rare audio}, which are smaller speech-based analogues of the original test sets. 40 dialogs corresponding to corpus-level distinct entity triplets were picked from each of the original test sets, and English-speaking human subjects of various ethnicities were asked to verbally read the dialogs with their own audio setup and record their audio, resulting in phonetically rich test sets. Two automated transcription systems (A and B) by Amazon were independently used to transcribe the collected audio, and each dialog transcription was aligned with the text of the original dialog based on edit distance followed by manual re-alignment to obtain the turn-level transcriptions. The Word Error Rate (WER) for the dialogs in \textit{test freq audio} and \textit{test rare audio} were in the range 0.1-0.39 by system A and in the range 0.08-0.35 by system B. For all experiments, we use the transcripts by system A.

\begin{figure}[htb]
  \centering
\includegraphics[width=\columnwidth]{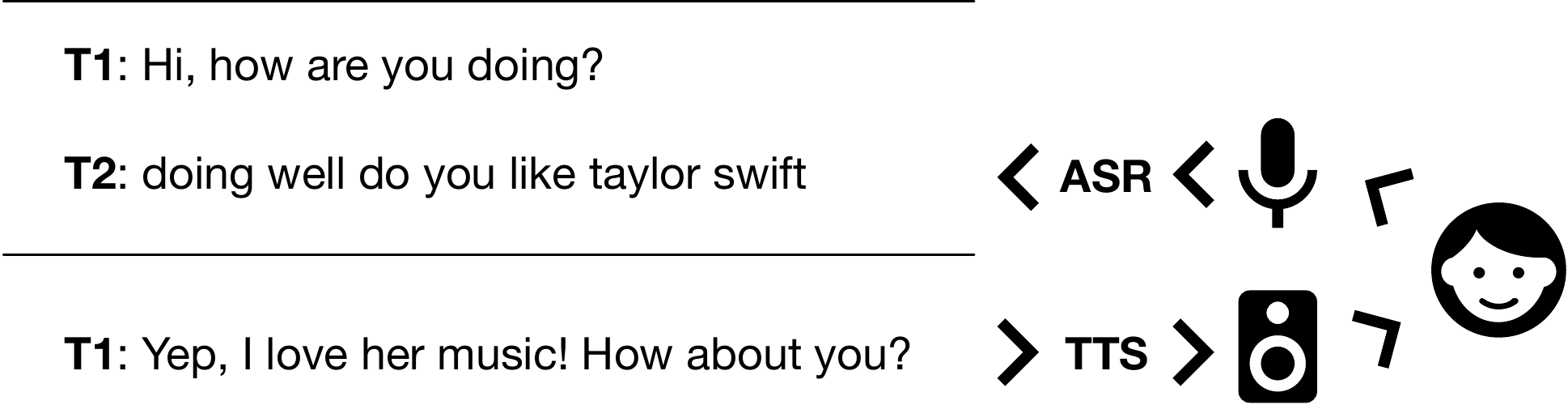}
\caption{\label{img-user-system}
\small A dialog example where the target turn is by Turker T1. We treat the Turker corresponding to the target turn (T1) as the \textit{system}, whose response is spoken to a user via a text-to-speech (TTS) engine. We treat the Turker corresponding to the previous turn (T2) as the \textit{user}, whose spoken utterance is converted into text via ASR. We use synthetic and actual ASR hypotheses for the user turns: in this example, we remove punctuation from T2's only turn in the dialog history.
}
\vspace{-0.15in}
\end{figure}

Since our focus is on studying the effect of synthetic and actual ASR hypotheses for user utterances, we need to label the partners in a dialog example as \textit{user} and \textit{system}. For a dialog history $D_t$ where turn $t+1$ is the target turn, we treat the Turker corresponding to turn $t+1$ as the \textit{system} and the
Turker corresponding to turn $t$ as the \textit{user}. This leads to system/user label assignments for all turns in $D_t$. Figure~\ref{img-user-system} shows this process.

\subsection{Models}

TransferTransfo~\cite{wolf2019transfertransfo} (TF2)
is a state-of-the-art neural open-domain dialog system by Hugging Face based on the Generative Pre-trained Transformer (GPT) model~\cite{radford2018improving} that won 1\textsuperscript{st} place in automated evaluation and 2\textsuperscript{nd} place in human evaluation at the NeurIPS ConvAI2 Conversational Intelligence
Challenge~\cite{dinan2020second}.

In this system, GPT is fine-tuned on a dialog dataset in a multi-task learning fashion with the \textit{language modeling} and \textit{next utterance classification} tasks. Starter code for TransferTransfo along with the pre-trained model and BPE-based
vocabulary is provided on GitHub by Hugging Face.

\textbf{Language modeling}: Trained with $(H_t^W, x_{t+1})$ pairs, this task is for learning to \textit{generate} a response given a dialog history.

\textbf{Next utterance classification}: This task is for learning to \textit{select} the ground-truth response $x_{t+1}$ from a set of $N$ candidate responses given a dialog history and is trained with $(H_t^W, \{c_{t+1}^i\})$ pairs, $i\in[1,\dots,N]$. The set of candidate responses contains distractors alongside the ground-truth. In our experiments, we used $N=5$ and randomly sampled turns from other dialogs in the same corpus (train/valid/test) as the one containing $D_t$ to serve as distractors.

To remove confounding factors that may affect our study of the effect of synthetic and actual ASR hypotheses in the dialog history $D_t$, we avoid conditioning TF2 on knowledge from the reading sets provided to Turkers in Topical-Chat.


\subsection{Training/Inference}
\label{trainingmarker}

Like~\cite{wolf2019transfertransfo}, we use special tokens to identify speakers with their associated segments and initialize them with random embeddings to be learned during fine-tuning. We fine-tuned for a fixed number of 3 epochs with equal weight of $1.0$ to the losses for both tasks. We wanted to fine-tune with a
large dialog history and set $W=256$. We used a train batch size of 2, performed gradient accumulation for 8 steps and gradient clipping with a max norm of $1.0$, used the Adam optimizer and linearly decayed the learning rate from 6.25e-5 to 0 during the course of training.
For simplicity, we set $H_t^W = x_t$ during inference time, i.e., we evaluate TF2's ability to generate and select given the last \textit{user} turn. We used top-$k$, top-$p$ nucleus sampling~\cite{holtzman2019curious} with temperature $T$
for decoding, where $k=0$, $p=0.9$ and $T=0.7$. We also set a maximum decode length of 40 tokens.



\section{Experimental Setup}
Unlike written text (which we denote by \textbf{GOLD}), raw ASR hypotheses typically do not contain punctuation and casing. Since TF2 performs lowercasing during tokenization, we exclude casing from consideration, and evaluate our models on the following three variations of the dialog history $D_t$ by using synthetic and actual ASR hypotheses:

\begin{itemize}
  \setlength\itemsep{0em}
  \item \textbf{NO-PUNC}: No punctuation in \textit{user} turns in $D_t$.
  \item \textbf{WER-SIM}: Simulated errors are introduced to \textit{user} turns in $D_t$ at a corpus-level target WER.
  \item \textbf{REAL}: Actual ASR errors are introduced to \textit{user} turns in $D_t$. This case is applicable for evaluation on \textit{test freq audio} and \textit{test rare audio} only.
\end{itemize}

We train these TF2 models as described in Section~\ref{trainingmarker}:
\begin{itemize}
  \setlength\itemsep{0em}
  \item \textbf{TF2-GOLD}: Trained on written text dialog examples.
  \item \textbf{TF2-NO-PUNC}: Trained on a NO-PUNC version of written text dialog examples.
  \item \textbf{TF2-WER-SIM}: For each WER, TF2 is trained on a WER-SIM version of written text dialog examples.
\end{itemize}

For the \textit{language modeling} task, we use the standard automated metrics of perplexity (PPL) and unigram F1 between the generated and ground-truth response.
For the \textit{next utterance classification} task, we use recall $R_N@k$~\cite{henderson2019training} with $N=5$ and $k=1$, which measures the accuracy of selecting the ground-truth response from a set of candidate responses.

\section{How Robust is TF2-GOLD?}

To answer this question, we evaluate TF2-GOLD in various settings and empirically compare them against the GOLD setting.

\subsection{NO-PUNC}
\label{gold-nopunc-eval}

Table~\ref{gpt-gold-nopunc} shows the results when evaluating TF2-GOLD in the NO-PUNC and GOLD settings on \textit{test freq} and \textit{test rare}. In order to better understand the
effect of sentence segmentation, we report results separately for three versions of each test set:

\begin{itemize}
  \setlength\itemsep{0em}
  \item \textit{single}: a subset of dialog examples where $H_t^W = x_t$ contains just a single sentence
  \item \textit{multi}: a subset of dialog examples where $H_t^W = x_t$ contains more than one sentence
  \item \textit{all}: \textit{single} $\cup$ \textit{multi}, i.e., all dialog examples
\end{itemize}



We observe a clear degradation in all metrics in the NO-PUNC setting relative to the GOLD setting, showing that TF2-GOLD relies on punctuation in the dialog history during both generation and selection. The degradation in PPL and $R_5@1$ when evaluating with the \textit{multi} version of the test sets is greater than when evaluating with the \textit{single} version, showing that the presence of sentence segmentation in dialog history comprised of multi-sentence utterances has a large impact on performance.



\begin{table}[ht]
\centering
\caption{\label{gpt-gold-nopunc}
\small TF2-GOLD evaluated in NO-PUNC and GOLD settings on the Topical-Chat test set (freq / rare).
}
\footnotesize
\begin{tabular}{cccc}
  \toprule
  \textbf{Metric} & \textbf{Version} & \textbf{NO-PUNC} & \textbf{GOLD} \\
  \midrule
  \multirow{3}{1.5cm}{\centering PPL}
  & {\textit{single}} & {18.0 / 23.9} & {17.9 / 23.6} \\
  & {\textit{multi}} & {19.2 / 25.3} & {18.8 / 24.7} \\
  & {\textit{all}} & {18.7 / 24.9} & {18.4 / 24.4} \\
  \midrule
  \multirow{3}{1.5cm}{\centering F1 (\%)}
  & {\textit{single}} & {16.6 / 16.2} & {16.4 / 16.4} \\
  & {\textit{multi}} & {15.9 / 15.5} & {16.0 / 15.6} \\
  & {\textit{all}} & {16.0 / 15.7} & {16.2 / 15.9} \\
  \midrule
  \multirow{3}{1.5cm}{\centering $R_5@1$ (\%)}
  & {\textit{single}} & {68.2 / 73.9} & {70.4 / 75.7} \\
  & {\textit{multi}} & {69.8 / 75.6} & {74.3 / 78.6} \\
  & {\textit{all}} & {69.7 / 74.8} & {73.7 / 77.7} \\
  \bottomrule
\end{tabular}
\end{table}

\subsection{WER-SIM}
\label{gold-wersim-eval}

Table~\ref{gpt-gold-wersim} shows the results when evaluating TF2-GOLD in the WER-SIM and GOLD settings on \textit{all} dialog examples of \textit{test freq} and \textit{test rare}. For each WER, we compute metrics using ASR hypotheses corresponding to all five seeds separately and report mean and standard deviation.
We observe a significant degradation in all metrics in the WER-SIM setting relative to the GOLD setting, showing that TF2-GOLD is very sensitive to simulated ASR hypotheses in the dialog
history. We also observe a significant degradation in metrics as the WER increases.
The magnitude of degradation in $R_5@1$ shows that response selection
relies heavily on the written form.


\begin{table}[ht]
\centering
    \caption{\label{gpt-gold-wersim}
    \small TF2-GOLD evaluated in WER-SIM and GOLD settings on the Topical-Chat test set (freq / rare). For each WER, the top and bottom rows contain the mean and standard deviation of the computed metrics using simulated ASR hypotheses corresponding to all five seeds.
    }
\footnotesize
\begin{tabular}{cccc}
  \toprule
  WER & PPL & F1 (\%) & $R_5@1$ (\%) \\
  \midrule
  \multirow{2}{*}{\centering 0.1}
  & {19.26 / 25.49} & {15.59 / 15.38} & {66.59 / 71.53} \\
  & {(0.007 / 0.005)} & {(0.08 / 0.04)} & {(0.18 / 0.01)} \\
  \midrule
  \multirow{2}{*}{\centering 0.15}
  & {19.55 / 25.81} & {15.42 / 15.32} & {65.08 / 70.13} \\
  & {(0.018 / 0.019)} & {(0.06 / 0.03)} & {(0.4 / 0.19)} \\
  \midrule
  \multirow{2}{*}{\centering 0.2}
  & {19.83 / 26.11} & {15.31 / 15.1} & {63.75 / 68.42} \\
  & {(0.01 / 0.02)} & {(0.09 / 0.06)} & {(0.2 / 0.35)} \\
  \midrule
  \multirow{2}{*}{\centering 0.3}
  & {20.46 / 26.83} & {14.89 / 14.77} & {59.96 / 64.47} \\
  & {(0.018 / 0.028)} & {(0.06 / 0.05)} & {(0.37 / 0.26)} \\
  \midrule
  \textbf{GOLD} & 18.39 / 24.42 & 16.17 / 15.95 & 73.67 / 77.74 \\
  \bottomrule
\end{tabular}
\end{table}


\subsection{REAL}

Table~\ref{gpt-gold-real} shows the results when evaluating TF2-GOLD in the REAL and GOLD settings on \textit{all} dialog examples of \textit{test freq audio} and \textit{test rare audio}. We observe a significant degradation in all metrics in the REAL setting relative to the GOLD setting, demonstrating that TF2-GOLD is very sensitive to actual ASR hypotheses in the dialog history.

\begin{table}[ht]
\centering
    \caption{\label{gpt-gold-real}
    \small TF2-GOLD evaluated in REAL and GOLD settings on our Topical-Chat audio test set (freq / rare).
    }
\footnotesize
\begin{tabular}{cccc}
  \toprule
 & PPL & F1 (\%) & $R_5@1$ (\%) \\
  \midrule
    \textbf{REAL} & 19.0 / 25.8 & 14.5 / 15.4 & 65.6 / 69.4 \\
    \midrule
    \textbf{GOLD}  &  17.4 / 24.1 & 15.5 / 16.1 & 76.3 / 78.6 \\
  \bottomrule
\end{tabular}

\vspace{-0.05in}
\end{table}

\section{Does Synthetic Training Help?}

We now study the efficacy of synthetic training in making TF2 robust to \textit{both} synthetic and actual ASR hypotheses.



\subsection{NO-PUNC}
We evaluate TF2-NO-PUNC in the NO-PUNC setting. Results for the NO-PUNC setting are in Table~\ref{gpt-nopunc} and can be interpreted jointly with Table~\ref{gpt-gold-nopunc}. We observe that TF2-NO-PUNC is more effective than TF2-GOLD at handling the NO-PUNC setting during inference time. There is a larger improvement in PPL and $R_5@1$ when evaluating with the \textit{multi} version of the test sets than the \textit{single} version, showing that NO-PUNC training is especially useful when the dialog history is comprised of multi-sentence utterances.

\begin{table}[ht]
\centering
\caption{\label{gpt-nopunc}
\small TF2-NO-PUNC evaluated in the NO-PUNC setting on the Topical-Chat test set (freq / rare).
}
\footnotesize
\begin{tabular}{cccc}
  \toprule
  & PPL & F1 (\%) & $R_5@1$ (\%) \\
   \midrule
   \textit{single} &  17.7 / 23.6 & 16.6 / 16.3 & 69.1 / 74.4 \\
   \midrule
   \textit{multi} & 18.7 / 24.7 & 16.0 / 15.6 & 72.8 / 76.9 \\
   \midrule
   \textit{all} & 18.3 / 24.4 & 16.2 / 15.9 & 70.9 / 76.4 \\
  \bottomrule
\end{tabular}

\end{table}

\vspace{-0.15in}
\subsection{WER-SIM}
We evaluate TF2-WER-SIM in the WER-SIM setting. Results for the WER-SIM setting are in Table~\ref{gpt-wersim} and can be interpreted jointly with Table~\ref{gpt-gold-wersim}. We observe that TF2-WER-SIM is more effective than TF2-GOLD at handling the WER-SIM setting during inference time.

\begin{table}[ht]
  \centering
\caption{\label{gpt-wersim}
\small TF2-WER-SIM evaluated in the WER-SIM setting on the Topical-Chat test set (freq / rare). Each row refers to a model trained with the corresponding WER: the top and bottom rows contain the mean and standard deviation of the computed metrics using ASR hypotheses corresponding to all five seeds.
}
\footnotesize
\begin{tabular}{cccc}
  \toprule
  WER & PPL & F1 (\%) & $R_5@1$ (\%) \\
  \midrule
  \multirow{2}{*}{\centering 0.1}
  & {18.69 / 24.92} & {16.19 / 15.94} & {69.77 / 74.32} \\
  & {(0.006 / 0.006)} & {(0.06 / 0.05)} & {(0.17 / 0.1)} \\
  \midrule
  \multirow{2}{*}{\centering 0.15}
  & {19.52 / 25.89} & {15.48 / 15.17} & {67.7 / 71.99} \\
  & {(0.018 / 0.013)} & {(0.07 / 0.07)} & {(0.04 / 0.2)} \\
  \midrule
  \multirow{2}{*}{\centering 0.2}
  & {19.36 / 25.8} & {15.75 / 15.48} & {66.74 / 71.45} \\
  & {(0.005 / 0.018)} & {(0.02 / 0.09)} & {(0.31 / 0.3)} \\
  \midrule
  \multirow{2}{*}{\centering 0.3}
  & {19.59 / 26.21} & {15.56 / 15.23} & {64.18 / 68.17} \\
  & {(0.009 / 0.015)} & {(0.11 / 0.08)} & {(0.28 / 0.15)} \\
  \bottomrule
\end{tabular}
\end{table}

\vspace{-0.15in}
\subsection{REAL: Automated Evaluation}
\label{realautomated}

We evaluate TF2-NO-PUNC and TF2-WER-SIM in the REAL setting. Results are in Table~\ref{gpt-wer-sim-real} and can be interpreted jointly with Table~\ref{gpt-gold-real}. We observe that TF2-NO-PUNC and TF2-WER-SIM are generally more effective than TF2-GOLD at handling the REAL setting during inference time.

\begin{table}[ht]
\centering
\caption{\label{gpt-wer-sim-real}
\small TF2-NO-PUNC and TF2-WER-SIM evaluated in the REAL setting on our Topical-Chat audio test set (freq / rare). 
}
\footnotesize
\begin{tabular}{cccc}
  \toprule
  \textbf{Model} & PPL & F1 (\%) & $R_5@1$ (\%) \\
  \midrule
  \multirow{1}{*}{\centering TF2-NO-PUNC}
  & {18.5 / 25.2} & {15.6 / 15.1} & {66.4 / 71.3} \\
  \midrule
  \multirow{1}{*}{\centering 0.1}
  & {18.31 / 25.27} & {15.64 / 15.65} & {66.93 / 71.09} \\
  \midrule
  \multirow{1}{*}{\centering 0.15}
  & {19.11 / 26.02} & {15.48 / 15.14} & {67.16 / 70.76} \\
  \midrule
  \multirow{1}{*}{\centering 0.2}
  & {18.77 / 25.73} & {15.11 / 15.49} & {65.68 / 71.09} \\
  \midrule
  \multirow{1}{*}{\centering 0.3}
  & {18.6 / 25.65} & {15.64 / 15.35} & {66.14 / 71.21} \\
  \bottomrule
\end{tabular}
\end{table}

Our evaluation shows that synthetic training provides reasonable improvements when the nature of errors matches during training and inference and very marginal improvements otherwise. An analysis of our \textit{simulated} and \textit{actual} ASR augmented data showed there are more insertion and deletion errors in the simulated data, and only about 10\% of substitution errors in \textit{actual} appear in \textit{simulated} ASR augmented data. But regardless of the existence of training-inference match/mismatch, the perfomance gap with TF2-GOLD in the GOLD setting (particularly prominent in $R_5@1$) demonstrates that there is still a need for creative error-specific and/or error-agnostic techniques to make monolithic neural open-domain dialog models \textit{demonstrably} robust to non-trivial target errors.

\vspace{-0.05in}
\subsection{REAL: Human Evaluation}

We also performed a human evaluation of the \textit{language modeling} task from the two tasks in TF2, specifically generating responses from TF2-GOLD, TF2-NO-PUNC and TF2-WER-SIM in the REAL setting on the Topical-Chat audio test sets. 100 unique snippets were prepared from each test set, where each snippet contained a \textit{written-text} dialog history and responses from all four models. But, the responses were obtained by feeding the models the \textit{error-distorted} version of the dialog history. For each test set, a pair of annotators were asked to annotate snippets 1-60 and 40-100 respectively, thus providing 20 overlapping annotations to compute inter-annotator agreement. The annotators were asked to annotate on a 5-point nominal scale (1: Strongly Disagree, 2: Disagree, 3: Neither Agree Nor Disagree, 4: Agree, 5: Strongly Agree) whether each response was appropriate (AP) for the provided \textit{written-text} dialog history.

The goal of this human evaluation was to see if there's any perceptible difference in responses generated from TF2-GOLD and the top 3 synthetic models with the least PPL in the REAL setting from Section~\ref{realautomated} when fed error-distorted dialog history.

\begin{table}[ht]
\centering
\caption{\label{human-eval}
\small Human evaluation of responses by TF2-GOLD, TF2-NO-PUNC and TF2-WER-SIM (0.1 and 0.3) in the REAL setting on our Topical-Chat audio test set (freq above, rare below). AP = appropriateness. We report mean and margin of error interval at 95\% confidence.
}
\footnotesize
\begin{tabular}{ccccc}
  \toprule
  \textbf{Model} & \textbf{GOLD} & \textbf{NO-PUNC} & \textbf{0.1} & \textbf{0.3} \\
  \midrule
  \multirow{2}{*}{\centering \textbf{AP}}
  & {3.2 $\pm$ 0.25} & {3.4 $\pm$ 0.26} & {3.2 $\pm$ 0.24} & {3.1 $\pm$ 0.25} \\ 
  & {3.6 $\pm$ 0.25} & {3.7 $\pm$ 0.23} & {3.7 $\pm$ 0.23} & {3.7 $\pm$ 0.23} \\ 
  \bottomrule
\end{tabular}
\end{table}

We bucketed ratings 1-2 and 4-5 when computing Fleiss' kappa annotator agreement and got scores of 0.54 and 0.38 on the two test sets. We observe in Table~\ref{human-eval} that TF2-NO-PUNC is marginally better than TF2-GOLD for both test sets and TF2-WER-SIM is marginally better than TF2-GOLD on one test set.


\section{Conclusion}

We empirically studied the effects of synthetic and actual ASR hypotheses in the dialog history on TF2, a large state-of-the-art text-based neural open-domain dialog system from the NeurIPS ConvAI2 challenge. We observed that TF2 trained on written data is very sensitive to such hypotheses introduced to the dialog history during inference time, demonstrating that text-based neural open-domain chatbots may not be very effective at serving the speech modality as-is. We observed that training TF2 with synthetic ASR hypotheses makes it more robust to both synthetic and actual ASR hypotheses during inference time, with considerable room for improvement left for future work. Our augmentation of Topical-Chat, dubbed the Topical-Chat ASR dataset, is open-sourced and we hope our work sparks discussion and further research into modality-centric and modality-agnostic open-domain dialog systems.

\bibliographystyle{IEEEtran}

\bibliography{asr}


\end{document}